\documentclass[10pt,twocolumn,letterpaper]{article}
\usepackage{cvpr}
\pdfoutput=1

\usepackage{etoolbox}
\usepackage{graphicx}
\usepackage{amsmath}
\usepackage{amssymb}
\usepackage{booktabs}

\usepackage{times}
\usepackage{epsfig}
\usepackage{algorithm}
\usepackage{algpseudocode}
\usepackage{array}
\usepackage{caption}
\usepackage{subcaption}
\usepackage[numbers]{natbib}
\usepackage{float}
\usepackage{enumitem}

\makeatletter
\robustify\@latex@warning@no@line
\makeatother
\usepackage{authblk}

\usepackage[pagebackref,breaklinks,colorlinks,bookmarks=false]{hyperref}

\def\ie{\textit{i.e.,}~}
\def\eg{\textit{e.g.,}~}

\DeclareMathOperator*{\argmin}{arg\,min}

\usepackage[capitalize]{cleveref}
\crefname{section}{Sec.}{Secs.}
\Crefname{section}{Section}{Sections}
\Crefname{table}{Table}{Tables}
\crefname{table}{Tab.}{Tabs.}

\def\cvprPaperID{6} 
\def\confName{EVW}
\def\confYear{2022}

\begin{document}

\title{\vspace{-3mm}Does Interference Exist When Training a Once-For-All Network?\vspace{-6mm}}  

\author[1]{Jordan Shipard}
\author[2]{Arnold Wiliem}
\author[1]{Clinton Fookes}

\affil[1]{\small Signal Processing, Artificial Intelligence and Vision Technologies (SAIVT), Queensland University of Technology, Australia}
\affil[2]{Sentient Vision Systems, Australia}
\affil[ ]{\textit {\tt \small \{jordan.shipard@hdr., c.fookes@\}qut.edu.au}, \textit{ \tt \small arnoldw@sentientvision.com}}

\maketitle
\thispagestyle{empty}


\begin{abstract}

The Once-For-All (OFA) method offers an excellent pathway to deploy a trained neural network model into multiple target platforms by utilising the supernet-subnet architecture. 
Once trained, a subnet can be derived from the supernet (both architecture and trained weights) and deployed directly to the target platform with little to no retraining or fine-tuning. 
To train the subnet population, OFA uses a novel training method called Progressive Shrinking (PS) which is designed to limit the negative impact of interference during training.
It is believed that higher interference during training results in lower subnet population accuracies.
In this work we take a second look at this interference effect. Surprisingly, we find that interference mitigation strategies do not have a large impact on the overall subnet population performance.
Instead, we find the subnet architecture selection bias during training to be a more important aspect.
To show this, we propose a simple-yet-effective method called Random Subnet Sampling (RSS), which does not have mitigation on the interference effect. 
Despite no mitigation, RSS is able to produce a better performing subnet population than PS in four small-to-medium-sized datasets; suggesting that the interference effect does not play a pivotal role in these datasets.
Due to its simplicity, RSS provides a $1.9\times$ reduction in training times compared to PS. A $6.1\times$ reduction can also be achieved with a reasonable drop in performance when the number of RSS training epochs are reduced. Code available at \url{https://github.com/Jordan-HS/RSS-Interference-CVPRW2022}

\end{abstract}
\vspace{-3mm}
\section{Introduction}

Deploying deep neural network models for real-world applications requires accurate and fast inference~\cite{xu_scaling_2018}. Generally, these accuracy and latency constrains are competing with one another, with only architecturally optimal networks being able to achieve both. Designing these optimal networks by hand is a challenging task and has led to the explosion of the neural architecture search (NAS) field.

Some initial NAS methods~\cite{baker_designing_2017,real_large-scale_2017,zoph_neural_2017} were based on reinforcement learning approaches and proved the concept viable. Unfortunately, these approaches required extremely high computational resources as they train hundreds of architectures. Moreover, these only searched the architectures which optimise accuracy.

The goal of the approach eventually shifted to producing optimal networks with high accuracy and fast inference. 
To this end, two different approaches were developed: (1) the direct NAS methods~\cite{cai_proxylessnas_2019, dai_fbnetv3_2021, hu_dsnas_2020, liu_darts_2019, wu_fbnet_2019}; and (2) the one-shot methods~\cite{bender_understanding_2018, brock_smash_2017, guo_single_2020, xia_progressive_2021}. The former constructs a continuous search space and utilises gradient descent for the search. Whilst, the later uses a discrete search space which only requires the neural network architecture search space to be trained once. This means the search becomes much faster and requires significantly fewer computational resources.

\begin{figure*}[t]

\centering
	\begin{subfigure}[Figure A]{\columnwidth}
		\includegraphics[scale=0.4]{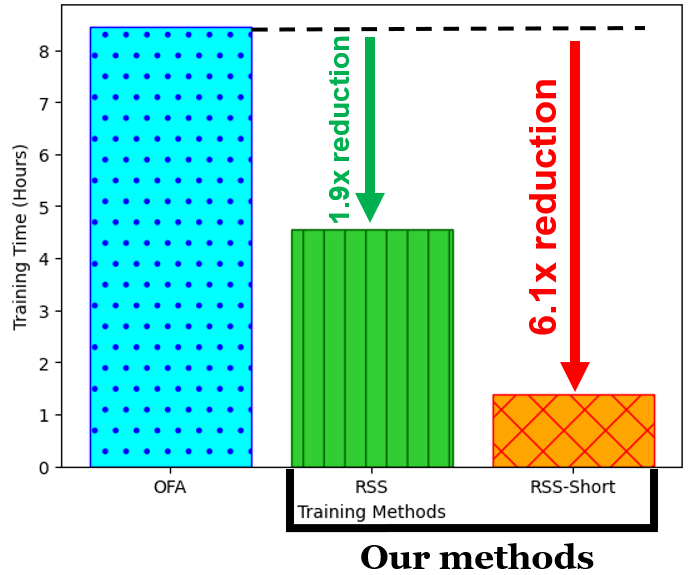}
		\captionsetup{width=8cm}
		\caption{Training time measured from the start of the first training epoch and the finish of the last epoch for OFA, RSS (proposed) and RSS-Short (proposed) methods. Refer to ~\cref{tab:training_epochs} for number of training epochs.}
	\end{subfigure}
	\begin{subfigure}[Figure B]{\columnwidth}
		\includegraphics[scale=0.4]{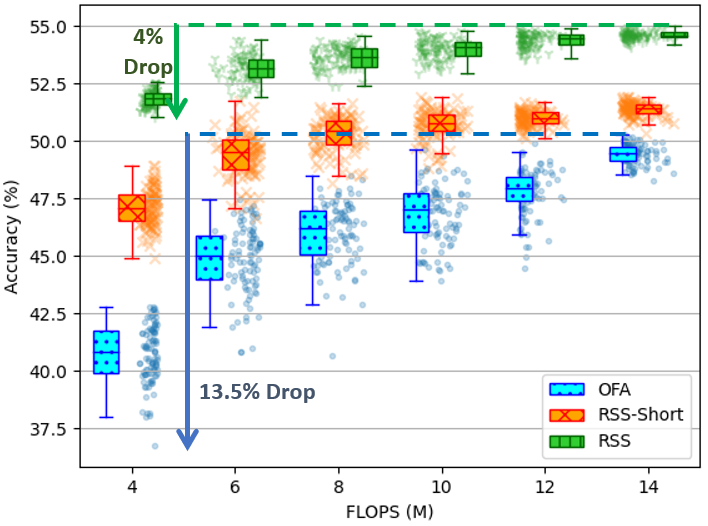}
		\captionsetup{width=8cm}
		\caption{Subnet population performance of the OFA, RSS (proposed) and RSS-Short (proposed). Both proposed variants have better top performing subnets with a lower variance in the population. More detailed results are reported in \Cref{sec:main_results}}
	\end{subfigure}

   \caption{Training time reduction and accuracy improvements of the proposed Random Subnet Sampling (RSS) compared to the Once-For-All (OFA) method. Results on CIFAR100 dataset ~\cite{krizhevsky_learning_2009}.}
\label{fig:short}
\end{figure*}

Previous one-shot methods~\cite{bender_understanding_2018, brock_smash_2017, guo_single_2020, xia_progressive_2021} use the trained search space as a guide for finding optimal networks. Once the optimal networks are found, they are trained from scratch before deployment. This presents a problem if we wish to deploy different network variants for various deployment platforms. To address this, the Once-For-All (OFA) method~\cite{cai_once-for-all_2020} trains a supernet wherein a sub-network/subnet architecture and its weights can be directly sampled from the supernet. Once sampled, a small amount of fine-tuning might be applied before the subnet is deployed; thus, side stepping the need for training the sampled subnets from scratch for each deployment platform.

To achieve its goal, OFA needs to train a large subnet population (\(2\times10^{19}\) subnets)~\cite{cai_once-for-all_2020}. When attempting to train the subnet population using naive approaches they find the subnets interfere with each other, resulting in significant accuracy drops. This is taken to suggest that modifying the weights of one subnet could affect the performance of other subnets in the subnet population. However, this interference is not currently well understood. To address the interference, OFA proposes a novel training method, called Progressive Shrinking (PS). PS trains the largest architecture first (\ie the supernet) and then progressively samples and trains smaller subnet architectures. Recent work in~\cite{sahni_compofa_2021} showed that it is possible to improve the accuracy of the subnets by reducing the search space. The work suggests the accuracy gained is due to a reduction in interference resulting from the reduced search space. Therefore, the limiting factor on improving subnet population performance appears to be related to the interference effect between subnets during training.

In this work, we take a second look at this effect. In particular we ask the following questions: (1) Does the interference effect exists? (2) If it exists, then by how much does it affect the subnet population's performance? (3) If any other factors impact the subnet population's performance? To examine these questions, this paper introduces a simple method dubbed Random Subnet Sampling (RSS), which randomly samples a single subnet to train at each epoch. Obviously, RSS does not have any mitigation on the interference effect. 
We compare RSS with OFA's PS method on four datasets: MNIST~\cite{lecun_mnist_2010}, Fashion-MNIST~\cite{xiao_fashion-mnist_2017}, CIFAR10~\cite{krizhevsky_learning_2009}, and CIFAR100~\cite{krizhevsky_learning_2009}.
To our surprise, the subnet population is able to better generalise and achieve higher accuracies than PS. 
Fig.~\ref{fig:short} shows the main findings from this work.

Our findings suggest that interference between subnets has a minimal effect on the subnet population's performance. 
Instead, we argue that bias in the subnet selection scheme during training has a larger impact on performance. 
When a subnet architecture is sampled and trained more often than the others, the subnet tends to have significantly higher accuracy. On the other hand, the performance of rarely sampled subnet architectures tend to have significantly lower accuracy.
This sampling bias is analogous to 
the bias introduced when training a neural network
model with an imbalanced dataset.
The proposed RSS method addresses this sampling bias by uniformly sampling the subnet architecture during each epoch.

We also observe that the interference effect becomes more apparent when combining multiple subnets gradients during a single update step. 
This corroborates recent findings by Xu~\etal~\cite{xu_analyzing_2021} in the Natural Language Processing (NLP) field. 	

\noindent
\textbf{Contributions - } Our contributions are listed as follows,
\begin{enumerate}
    \item In contrast to the recent belief, we show that the interference has minimal effect when training the subnet population. 
    \item Instead, we argue that bias in the subnet selection scheme during training plays a bigger role.
    \item We also show that the interference effect becomes more pronounced when combining the gradients of multiple subnets in a single update step.
    \item To show the above points, we propose a simple-yet-effective method called Random Subnet Sampling (RSS). The proposed RSS method outperforms Once-For-All's Progressive Shrinking (PS) method which suggests point (1). In addition, the reason why RSS has good performance is because it addresses the bias problem as stated in point (2) and it only trains a single subnet for each epoch, in line with point (3).
\end{enumerate}

We continue our paper as follows. 
Related works are discussed in~\Cref{sec:related_work}. 
The subnet population training problem formulation is presented in~\Cref{sec:problem_definition}.
We then introduce the proposed method in~\Cref{sec:proposed_method} and discuss subnet sampling bias during training in~\Cref{sec:subnet_sampling_bias}.
~\Cref{sec:results} presents the experimental results. Finally, ~\Cref{sec:conclusion} discusses conclusions and the future direction of this work.

\section{Related Work}
\label{sec:related_work}


\noindent 
\textbf{One-Shot NAS -}
The goal of Neural Architecture Search (NAS) is to search for an optimal architecture in a large architecture search space. This search space can be continuous, in the case of direct NAS methods~\cite{cai_proxylessnas_2019, dai_fbnetv3_2021,hu_dsnas_2020, liu_darts_2019, wu_fbnet_2019}, or discrete, in the case of one-shot methods~\cite{bender_understanding_2018, brock_smash_2017, guo_single_2020}. Additionally, the name `one-shot` refers to the subnet population only requiring to be trained once; whereas previous methods used reinforcement learning and trained hundreds of individual networks during search~\cite{baker_designing_2017,real_large-scale_2017,zoph_neural_2017}. One-shot methods, such as Once-For-All~\cite{cai_once-for-all_2020}, train large discrete search spaces using weight sharing techniques~\cite{chen_net2net_2016, ha_hypernetworks_2016, han_learning_2015, yu_slimmable_2018} removing the requirement to train every architecture.  The relative accuracy of subnets in the subnet population is then used to find an optimal architecture. Once found, the architecture is trained from scratch as the previous shared weights are not fully optimised. As a further improvement, various methods were able to remove this need for retraining~\cite{cai_once-for-all_2020, huang_ponas_2020, liu_progressive_2018, sahni_compofa_2021, wang_attentivenas_2021, yu_bignas_2020}, allowing subnets to be directly extracted from the subnet population. Despite its computational benefits, weight sharing is not a perfect solution as it is believed to introduce the problem of interference between subnets. In this work we look to further study this interference as it is currently under explored.

\noindent
\textbf{Interference Effect in One-Shot Training -}
Most of the current understanding around the effects of interference come from observations made during the application of specific methods. Guo \etal~\cite{guo_single_2020} states that the weights in the supernet are deeply coupled, although no further study to explain the underlying mechanism of coupling is provided.
Cai \etal~\cite{cai_once-for-all_2020} state that randomly selecting subnets for training causes interference and accuracy drops; which Sahni \etal~\cite{sahni_compofa_2021} echo and believe their compound heuristic is able to reduce interference and therefore improve accuracy. Liu \etal~\cite{xia_progressive_2021} state that the existence of unnecessary neurons and connections in the supernet negatively impacts training and leads to greater interference. 
Perhaps, the most closely related study of interference is from the field of Natural Language Processing (NLP) by Xu \etal~\cite{xu_analyzing_2021}. They conduct a specific analysis of the interference effect during training a single-path one-shot NAS method.
Their work finds that interference between subnets is caused by diverse gradient directions produced by multiple sampled subnets. Combining these diverse gradient directions would hamper the training progress.
Furthermore, large architectural differences between sampled subnets contribute to the diverse gradient directions.
Our work differs as we train a subnet population constructed from a computer vision-based neural network model, instead of a single-path subnet population for NLP. Despite these differences we find their findings to be consistent with our own. 

\noindent
\textbf{Bias in One-Shot Training -}
The effect of bias during one-shot training has been raised recently by Chu \etal~\cite{Chu_fairnas_2021}.
Their work observe an unfair bias in the previous one-shot methods'~\cite{pham_efficient_2018, liu_darts_2019} supernet training as the cause for the poor correlation between the subnet proxy performance and its corresponding standalone trained performance (\ie trained from scratch).
Since the proxy performance is used as a guide when searching for optimal architectures, low correlation with the standalone trained performance will produce suboptimal architectures.
To this end, they propose a strict fairness based training method. Despite bias being the core issue, their work only focus on the training fairness issue and does not extensively explore the bias itself. 
Different to their work, in this work we explain bias in the subnet sampling/selection during training of a OFA subnet population, to be analogous to bias in the data imbalance problem.
This analogy allows us to borrow solutions developed for this problem to help us address the subnet sampling bias.






 
\noindent
\textbf{Data Imbalance - }
Real world data is not balanced~\cite{weiss_effect_2001} like the ones commonly used in scientific works. This data imbalance or bias is a well known problem within image classification and many solutions have been developed to address it. These include; Balanced sampling \cite{estabrooks_multiple_2004, laurikkala_improving_2001, weiss_effect_2001}; Hard mining \cite{loshchilov_online_2016, shrivastava_training_2016, simo-serra_fracking_2015, wang_unsupervised_2015} which directly relate to ideas used by Wang \etal~\cite{wang_attentivenas_2021} for worst-up training; and focal loss \cite{dong_focal_2020, lin_focal_2018}. 
As mentioned, we propose that the subnet sampling bias can be explained from the data imbalance perspective.


\section{Problem Description}
\label{sec:problem_definition}

We follow the problem definition presented in the original OFA work~\cite{cai_once-for-all_2020}. 
Let $\operatorname{arch}_i \in \mathcal{A}$ be the $i$-th subnet architecture. 
It is assumed that all subnet architectures in $\mathcal{A}$ can be derived/sampled from the supernet (the largest model amongst all).
Let $\mathbf{W_o}$ be the weights of the supernet and $\mathbf{W_{\operatorname{arch}_i}}$ be the weights of $\operatorname{arch}_i$. 
We derive $\mathbf{W_{{\operatorname{arch}_i}}}$ from $\mathbf{W_o}$ via a selection function $\operatorname{C} ( \cdot )$. 
The subnet population training problem can then be formalised via a minimisation problem as,
\begin{align} 
\label{eq:problem_definition}
    \argmin_{\mathbf{W_o}} \sum_{arch_{i} \in \mathcal{A}} \mathcal{L}_{val}(C(\mathbf{W_o}, arch_i)), 
\end{align}
where $\mathcal{L}_{val}$ is the loss on the validation set.
Essentially, the above equation aims at minimising the loss of every subnet from $\mathcal{A}$ on the validation set. In the case of the OFA, and for this paper, the size of $\mathcal{A}$ is $2 \times 10^{19}$ subnets.



\section{Subnet Population Training}
\label{sec:proposed_method}

Due to the large population size (\ie $|\mathcal{A}| = 2 \times 10^{19}$), directly minimising Eq.~\ref{eq:problem_definition} is impractical. The general approach is to sample and train a subset $\tilde{\mathcal{A}} \in \mathcal{A}$.
The subset $\tilde{\mathcal{A}}$ is sampled according to some selection scheme and set via $\operatorname{C} ( \cdot )$. 
Any subnet that does not belong to $\tilde{\mathcal{A}}$ will still be indirectly trained due to its weights being shared with the subnets in $\tilde{\mathcal{A}}$. 
In this section we first briefly discuss PS, proposed in the original OFA paper~\cite{cai_once-for-all_2020}, and then propose Random Subnet Sampling (RSS).

\subsection{Progressive Shrinking}
\label{sec:PS}
The main idea of the PS~\cite{cai_once-for-all_2020} is to train the subnets with respect to their Floating Point Operation (FLOP) size; from the largest one (\ie supernet) to the smaller ones progressively. 

The supernet is required to be trained prior to using the progressive shrinking method. Once the supernet is trained, PS trains the subnet population in three stages by performing a controlled sampling on three network parameters: (1) kernel size (\eg 7x7, 3x3); (2) number of layers in each block (Depth); and (3) number of channels (Width). 

In the first phase, named dynamic kernel training, only the kernel size is varied while the depth and width are kept at their maximum values. In this stage a single subnet is sampled for each update step. In the second phase, called dynamic depth training, the method varies both the number of layers and the kernel size, while width still remains at its maximum value. This stage samples two subnets and combines the gradient from both for each update step. Finally, in dynamic width training, the method varies all three parameters and combines the gradients from four sampled subnets during each training step.

\begin{algorithm}[b]
\caption{Pseudo code for the proposed Random Subnet Sampling (RSS) method. Kernel\_settings, width\_settings and depth\_settings are sets of values. For instance, width\_settings = \{3, 4, 6\}. $\operatorname{rand} ( \cdot )$ is a sampling function that uniformly samples values from the input argument.}
\label{alg:training}
\begin{flushleft}

\hspace*{\algorithmicindent} \textbf{Input:} kernel\_settings, width\_settings, depth\_settings, \\ \hspace*{\algorithmicindent}  n\_epochs \\
\hspace*{\algorithmicindent} \textbf{Output:} Trained subnet population
\begin{algorithmic}[1]

\While {$i\leq n\_epochs$}
	\State$\mathbf{subnet\_k} \gets \operatorname{rand} (kernel_{settings})$
	\State$\mathbf{subnet\_e} \gets \operatorname{rand} (width_{settings})$
	\State$\mathbf{subnet\_d} \gets \operatorname{rand} (depth_{settings})$
    \State$subnet \gets \operatorname{C} ( \mathbf{subnet\_k}, \mathbf{subnet\_e}, \mathbf{subnet\_d})$
    \State train $subnet$ for one epoch
\EndWhile

\end{algorithmic}
\end{flushleft}
\end{algorithm}

\subsection{Random Subnet Sampling}

We propose Random Subnet Sampling (RSS) which is not designed to mitigate the interference effect. 
In contrast to the progressive shrinking method, RSS does not perform any controlled sampling. It samples the subnets by varying all three network parameters used in PS (\ie the kernel size, depth and width). We use uniform randomness to choose the value for each parameter.
Unlike PS which samples subnets for each update step, or batch, RSS samples a single subnet for each epoch.
We show later that per-epoch sampling is a more effective method than per-batch sampling.
Algorithm~\ref{alg:training} illustrates the proposed RSS method. 
The kernel, width  and depth settings are one dimensional vectors with the length prescribed as follows. 

The length of the kernel settings vector ($\mathbf{subnet\_k}$) and width settings vector ($\mathbf{subnet\_e}$) is equal to the maximum depth setting multiplied by the number of blocks in the supernet. As an example, if the supernet has five blocks, with the maximum depth setting being four. Our kernel and width settings vectors would therefore contain 20 values (i.e. 5 blocks $\times$ 4 layers = 20). The length of the depth settings vector ($\mathbf{subnet\_d}$) is equal to the number of blocks. As we see from Algorithm \ref{alg:training}, the values of these three setting vectors are randomised between lines 2-3. The subnet is then derived by passing the sampled settings into the selection function $\operatorname{C}(\cdot)$ and trained.
More specifically, we follow OFA~\cite{cai_once-for-all_2020} to derive the subnet and its weights from the supernet by using the sampled settings.




\section{Subnet Sampling Bias} 
\label{sec:subnet_sampling_bias}

Different sampling strategies between PS and the proposed RSS can be studied from the sampling bias perspective.
We argue that Eq.~\ref{eq:problem_definition}, bears similarities to  the standard classification problem~\cite{lecun_gradient_1998}.
To train a classification model, we can solve the following minimisation problem.
\begin{align} 
\label{eq:classification_problem}
    \min_{\mathbf{W^*}}\sum_{i=1}^M \mathcal{L}_{train}(\mathbf{W^*}, \mathbf{x_i}, y_i),
\end{align}
Where $\mathbf{x_i}$ and $y_i$ are the $i$-th data point and its corresponding class label. $\mathbf{W^*}$ is the neural network weights and $\mathcal{L}_{train}$ is the training loss. 

Similar to Eq.~\ref{eq:problem_definition}, the above minimisation problem aims at reducing the loss for each data point in the training set, $( \mathbf{x_i}, y_i)_{i=1}^{M}$.
The difference for Eq.~\ref{eq:problem_definition} is that instead of
summing the loss over all the data points, it sums the loss over all the subnet architectures.

Linking the subnet population training problem presented in Eq.~\ref{eq:problem_definition} with the classification problem in Eq.~\ref{eq:classification_problem} is appealing as we can use the tools/experience developed in the classification problem into this field. For instance, bias in the data is one of the big themes in the classification field~\cite{weiss_effect_2001}. When training a network with an imbalanced training set (\ie some classes have significant higher number of training data than others), the network will tend to have stronger confidence scores towards these large classes. 

Using this new perspective, we argue that PS is a biased selection scheme, as it initially trains the supernet before training the largest to smallest subnets. As described in~\Cref{sec:PS}, PS has three phases which progressively varies the three network parameters. As an example, during dynamic kernel training the width and depth are set at their maximum values. This results in only larger subnets being trained during this stage.
Overall this could lead to the effect obsereved by Chu \etal~\cite{Chu_fairnas_2021}, with the larger subnets (especially the supernet) performing better than smaller subnets, simply due to more training.

Unlike PS, RSS randomly samples subnets for the entire training duration. This means, the subnet sampling is not biased towards the larger subnets.
We will show in our results that the subnet sampling bias does indeed play a significant role in the overall performance; and that this role is similar to the data sampling bias in the classification problem.

\section{Experimental Results}
\label{sec:results}
We contrast between the proposed RSS method and PS in this section. If the interference effect has significant contribution, then RSS will have significant lower performance than PS. First, the experimental set-up and the datasets are discussed. Then, we present our main results. 
Finally, we show additional ablation experiments.
%
%
%
%
%
\subsection{Experiment Setup}
\noindent 
\textbf{Datasets and Methods - } 
The datasets used for comparison are MNIST ~\cite{lecun_mnist_2010}, Fashion-MNIST (FMNIST)~\cite{xiao_fashion-mnist_2017}, CIFAR10~\cite{krizhevsky_learning_2009}, CIFAR100~\cite{krizhevsky_learning_2009}.
%
%
%
%
These datasets were used to cover a range of dataset complexities with MNIST being the least complex and CIFAR100 being the most complex. 

 \begin{table}[b]
    \centering
    \begin{tabular}{||c|c|c|c||}
    \hline
        &\multicolumn{3}{|c||}{Training Epochs} \\
        \hline
         Dataset&OFA&RSS-Short&RSS \\
         \hline
         \hline
         MNIST&10(s)+27(ps)&10&37 \\
         F-MNIST&25(s)+57(ps)&25&82 \\
         CIFAR10&180(s)+410(ps)&180&590 \\
         CIFAR100&180(s)+410(ps)&180&590 \\
         \hline
    \end{tabular}
    \caption{Training epochs for each method on each dataset where (s) denotes supernet training and (ps) denotes progressive shrinking training. For CIFAR10 and CIFAR100 we use the same training protocol as detailed by Cai \etal~\cite{cai_once-for-all_2020}. Training on MNIST and F-MNIST are scaled versions of the same CIFAR10 and CIFAR100 training with less epochs.}
    \label{tab:training_epochs}
\end{table}

We compare three training methods as follows.

\noindent
\textbf{OFA~\cite{cai_once-for-all_2020}-} The progressive shrinking method is used.

\noindent
\textbf{RSS-} The proposed random subnet sampling with the same number of training epochs as OFA.

\noindent
\textbf{RSS-short-} The proposed RSS with the same number of training epochs as only OFA's supernet training.The training epochs for each method are shown in \cref{tab:training_epochs}.

All training was conducted on a single Nvidia RTX3080 Graphics Processing Unit (GPU) with a initial learning rate of 0.01 using cosine learning rate decay; batch size of 64; momentum of 0.9; weight decay of $3e^{-5}$ and used cross-entropy loss, unless otherwise specified.

\begin{figure}[b]
	\centering
	\includegraphics[scale=0.5]{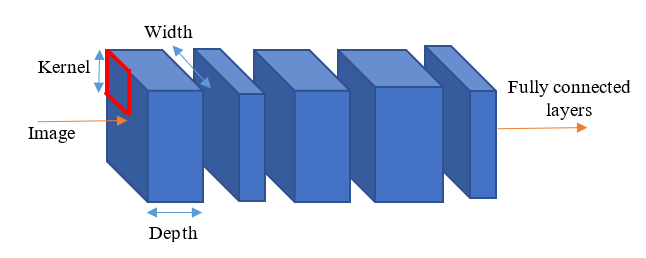}
	\caption{Diagram showing the MobileNetV3~\cite{howard_searching_2019} base architecture used to sample subnets with kernel $\in$ \{3x3,5x5,7x7\}, width $\in$ \{3,4,6\} and depth $\in$ \{2,3,4\}.}
	\label{fig:ofa_net}
\end{figure}

\begin{figure*}[t]
\centering
\captionsetup{justification=centering}
\begin{subfigure}[Figure A]{\columnwidth}
         \centering
         \includegraphics[scale=0.55]{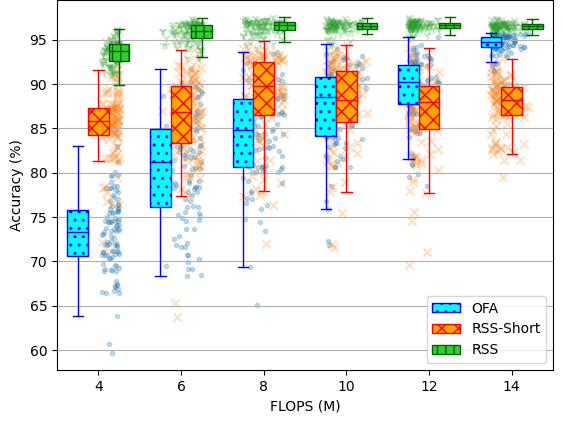}
         \caption{MNIST Subnet Population Performance.}
         \label{fig:y equals x}
     \end{subfigure}
     \hfill
     \begin{subfigure}[Figure B]{\columnwidth}
         \centering
         \includegraphics[scale=0.55]{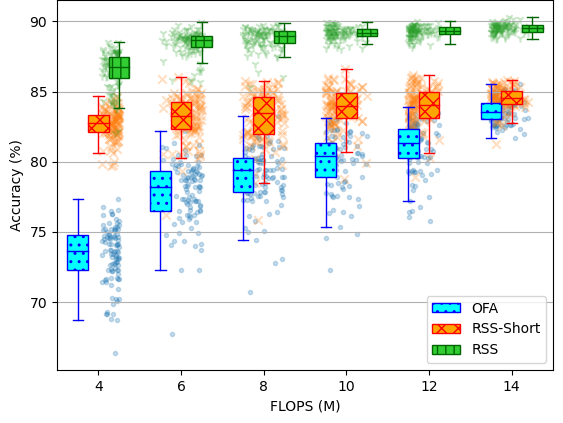}
         \caption{FMNIST Subnet Population Performance.}
         \label{fig:three sin x}
     \end{subfigure}
     \hfill
     \begin{subfigure}[Figure C]{\columnwidth}
         \centering
        \includegraphics[scale=0.55]{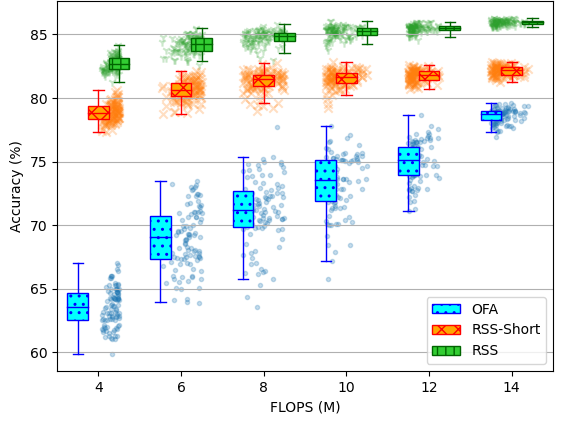}
        \caption{CIFAR10 Subnet Population Performance.}
        \label{fig:my_label}
     \end{subfigure}
     \hfill
     \begin{subfigure}[Figure D]{\columnwidth}
        \centering
        \includegraphics[scale=0.55]{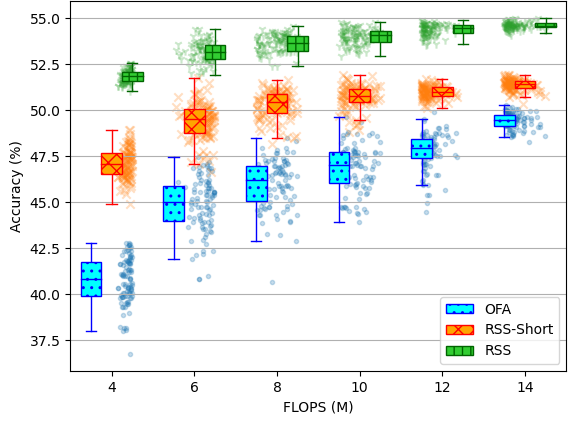}
        \caption{CIFAR100 Subnet Population Performance.}
        \label{fig:my_label}
     \end{subfigure}
     
     \caption{Comparisons between OFA, RSS (proposed), and RSS-Short (proposed) subnet population performance across four datasets. RSS is able to consistently train a better performing subnet population for all datasets.}
     \label{fig:main_results}
\end{figure*}
\noindent 
\textbf{Subnet Population Architecture - }
In this work, the MobileNetV3~\cite{howard_searching_2019} is used as the base architecture. From our empirical observations (provided in supplementary material), our findings are still consistent when a different base architecture is used.

The subnet derivation from the base architecture is kept consistent with the settings used by OFA.
The subnet population is defined along three mutable dimensions, the kernel size, layer width (number of channels), and layer depth (number of layers for each block). The possible settings for each dimension are shown below.
\begin{itemize}[itemsep=0.11pt,topsep=0pt]
    \item Kernel $\in \{3\times3, 5\times5, 7\times7\}$.
    \item Width $\in \{3, 4, 6\}$.
    \item Depth $\in \{2, 3, 4\}$.
\end{itemize}

The supernet consists of five configurable blocks, shown in \cref{fig:ofa_net}. All subnets are unique combinations of the kernel, width and depth settings. The kernel is set layer by layer and controls the active kernel size. The width, or expansion ratio, is also set layer by layer and is a multiplier for the base channel count of each layer. The depth is set for each block and controls the number of active layers in that block. We follow OFA~\cite{cai_once-for-all_2020} for transforming supernet weights to derive the prescribed subnet.

\noindent
\textbf{Evaluation Protocol - } OFA~\cite{cai_once-for-all_2020} evaluate their training methods performance according to the top performing subnets at various latency constraints. However, improving the performance of top-performing subnets may not correlate to a better performing subnet population. We instead wish to evaluate the average performance of the overall subnet population.
This is done by randomly sampling a group of subnets according to the specific architectural size measurement of Mega FLoating-point Operation Per Seconds (MFLOPs).
The performance of all sampled subnets is then recorded on the test set.
For all datasets, we use the following architecture sizes of evenly spaced MFLOPs values $\{4, 6, 8, 10, 12, 14\}$. 
Subnets of size 6-12 MFLOPs can be found by simply randomising all settings until the resulting subnet is within the required MFLOP range, $\pm0.5$ in our case. 
Subnets with size 4 and 14 MFLOPs are rare and challenging to 
sample in this manner.
As such, to find these subnets, the width setting is locked at its minimum value to aid sampling of 4 MFLOP subnets; and locked at its maximum to aid sampling of 14 MFLOP subnets.
We use the wall clock to measure the training time for each method from the start of the first epoch to the conclusion of the last epoch.

\subsection{Main Results}
\label{sec:main_results}
\noindent 
\textbf{Subnet Population Performance -} 
The main results are presented in \cref{fig:main_results}.
Across all datasets, RSS achieves the best performing subnet population.
The gap between RSS and OFA grows larger as the difficulty of the dataset increases (i.e., from MNIST to CIFAR).
Furthermore, the range of best performing to worst performing subnets is significantly smaller for RSS methods. 
These results suggest that the interference effect between subnets during training is negligible. 

In \Cref{sec:RSS_outperform} we show RSS's consistent performance across different subnet sizes is due to both its unbiased subnet sampling scheme and reduction of interference.
In contrast to RSS, OFA shows a significant bias towards larger subnets in all datasets.
We attribute this to its initial supernet training and PS training procedure which starts from the larger subnets.
Subnet sampling bias is further studied in \Cref{sec:selection_scheme}.

\begin{figure}[t]
    \centering
    \includegraphics[scale=0.55]{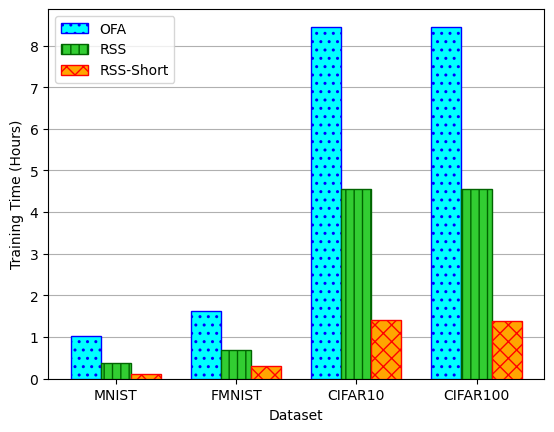}
    \caption{Comparisons of the training time, in hours, for each method as measured on a single Nvidia RTX3080 GPU.}
    \label{fig:training_speed}
\end{figure}

\noindent 
\textbf{Training Time -} 
\cref{fig:training_speed} shows the number of hours required to train each of the methods. Showing RSS and RSS-short are an average 1.9 and 6.1 times faster than OFA respectively. For RSS-short this is expected as it runs for nearly half the epochs of OFA; however, RSS and OFA run for the same number of epochs. This speed up is due to two factors: (1) RSS is more likely to train smaller subnets each epoch; and (2) RSS only trains a single subnet each epoch where OFA's PS trains two subnets during dynamic depth training and four during dynamic width training.

\subsection{Why does RSS outperform OFA?}
\label{sec:RSS_outperform}
Our main results suggest that both proposed RSS variants outperform OFA. 
In this section we conduct further investigations by studying the two main differences between the methods: (1) the number and frequency of subnets sampled during training; and (2) the selection scheme for selecting subnets during training. 
We only present the results from CIFAR100 dataset for the study.
We choose CIFAR100 as it offers more variations and complexity compared to the other datasets.
Furthermore, we only use RSS with the standard number of training epochs (590). Results (in supplementary material) on RSS-Short are similar; in addition to showing our results are not limited to specific hyperparemeter values.

\vspace{-3mm}
\subsubsection{Effect of Number of Subnets Sampled}
\label{sec:subnet_sampled}
\begin{figure}[t]
	\centering
	\includegraphics[scale=0.5]{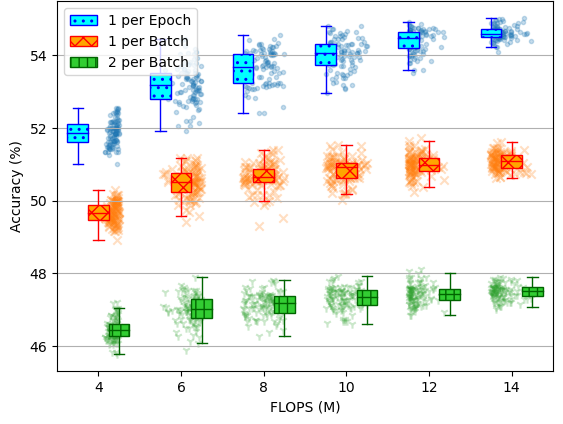}
	\caption{Effect of sampling different numbers of subnets during training. The 1 per epoch strategy is the default RSS method. When sampling 2 per batch the gradients are combined for the update step. As we can see, performance drops for per-batch strategies. The drop is more significant when more subnets are sampled per-batch (i.e., 2 per batch).}
	\label{fig:sample_rate_change}
\end{figure}

The proposed RSS method samples a single subnet to train each epoch, whereas OFA samples and trains up to four subnets per update step.
In this section, we modify the RSS subnet sampling to be similar to OFA, allowing us to compare subnet sampling methods on RSS.
More specifically, we first change RSS's sampling from each epoch into sampling each batch (i.e., per update step).
We then increase the number of sampled subnets for each batch from one to two.
When sampling two or more subnets, we combine their gradients in each update step.
\cref{fig:sample_rate_change} shows that despite training fewer subnets, per-epoch sampling produces a better performing subnet population than per-batch sampling. 
Indicating interference may be more severe in per-batch strategies. As an alternative explanation, we speculate there could be a benefit to allowing sampled subnets to train on the entire dataset instead of only a single batch. Future study is required to investigate this.
The interference again appears to increase between sampling one and two subnets per batch. We conjecture this increased interference is likely due to the combining of diverse gradient directions, as also shown in~\cite{xu_analyzing_2021}.

\vspace{-3mm}
\subsubsection{Effect of Different Selection Schemes}
\label{sec:selection_scheme}

The previous section investigated the number of subnets to sample. In this section, we study the subnet selection scheme.
As mentioned in \Cref{sec:subnet_sampling_bias}, the selection scheme can induce bias which skews the subnet population's performance towards the more frequently sampled subnets. In this section, we intentionally use biased subnet selection schemes to study the effects on the subnet population.

\begin{figure}
	\centering
	\includegraphics[scale=0.5]{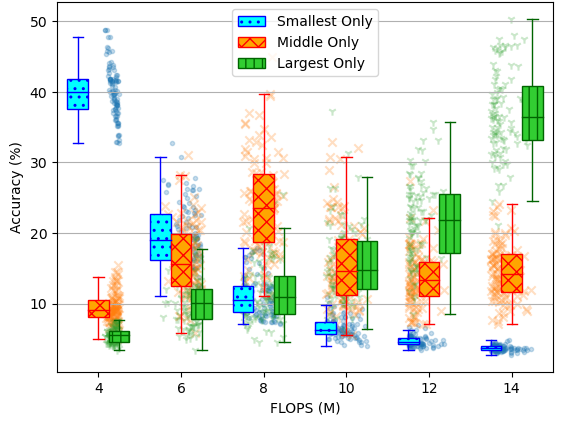}
	\caption{Resulting population performance from training a single subnet only. The \textbf{smallest only} uses only the smallest subnet which is constructed by settings all settings to their minimum. The \textbf{middle only} is when kernel=5, expand=4 and depth=3. The \textbf{largest only} is the supernet, when all settings are at their maximum. The results show a heavy bias towards the selected subnet.}
	\label{fig:anchor}
\end{figure}

\noindent
\textbf{Single subnet selection - } The simplest biased subnet selection scheme to test is training a single subnet.
More specifically, we compare three variants of this biased selection scheme: (1) \textbf{smallest only}, with all settings at their minimum values; (2) \textbf{middle only}, where kernel=5, expand=4 and depth=3 for all blocks; and (3) \textbf{largest only}, where all settings are set to their maximum (i.e. the supernet). 
\cref{fig:anchor} shows that the subnet performance is skewed towards the size of the selected subnet. Furthermore, the performance drops off as soon as the subnet size deviates from the selected subnet size.
Indicating excessive sampling of a subnet during training induces a strong bias in the subnet population's performance.

\begin{figure}
	\centering
	\includegraphics[scale=0.5]{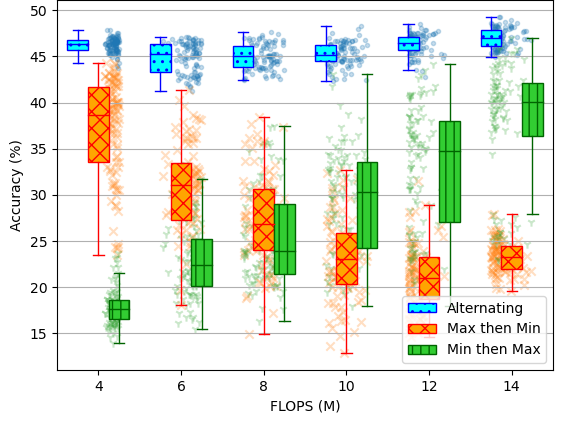}
	\caption{Resulting population performance from training the same two largest and smallest subnets in differing sequences, showing that training sequence does matter.}
	\label{fig:anchor_seq}
\end{figure}

\noindent
\textbf{Two subnet selection - } 
\cref{fig:anchor} shows that single subnet selection introduces a bias towards the selected subnet.
In this part, we examine the effect of training two subnets.
More specifically, we select either the smallest or the largest subnet for training.
We compare three variants of selection schemes: (1) select the largest subnet for the first half of the total epochs, then select the smallest subnet the rest of the training (\textbf{Max then Min}); (2) select the smallest subnet for the first half of the total epochs, then select the largest subnet the rest of the training (\textbf{Min then Max}); and (3) alternating between smallest and largest subnets each epoch (\textbf{Alternating}).
Unlike OFA which uses cosine learning rate decay, in this part of experiment, we use a constant learning rate throughout the training course.
We report the results in \cref{fig:anchor_seq}.

Both \textbf{Min then Max} and \textbf{Max then Min} schemes skew their results towards the subnet most recently trained; however, it is clear that both subnet populations benefited from the training of an additional subnet. This is shown by \textbf{Max then Min} having better performing subnets at larger MFLOPs when compared to training the smallest subnet only in \cref{fig:anchor}. The same can be observed for \textbf{Min then Max} at lower MFLOPs when compared to training the largest subnet only in \cref{fig:anchor}.
Interestingly, we do not see mirrored results between \textbf{Min then Max} and \textbf{Max then Min} schemes.
This shows that the training sequence within the selection scheme is an important factor. 

Another interesting observation is that the \textbf{Max then Min} scheme has lower performance than the \textbf{Min then Max} scheme.
OFA has a similar strategy to the \textbf{Max then Min}. However, OFA uses a decreasing learning rate and a more controlled selection scheme which might help to mitigate this issue.
A future study on this is warranted.

The \textbf{Alternating} scheme produces the best results with a more evenly trained subnet population and only slight bias towards the largest and smallest subnets.
This further suggests that mitigating the subnet sampling bias during training is an important issue to address.
Interestingly, despite never being trained, the subnets in ranges 8-10 MFLOPs perform nearly on par with the larger and smaller subnets. 
The \textbf{Alternating} scheme uses an alternating selection strategy, producing a less biased subnet population towards both trained subnet sizes. 
The proposed RSS is a step further by enlarging the subnet selection from two subnets to the whole subnet population. Moreover, RSS uses random sampling which enables each individual subnet to have the same chance to be selected for training. The results show there is benefit to sampling and training more subnets as RSS produces a better performing subnet population than the \textbf{Alternating} scheme and the other methods. 

\section{Conclusion}
\label{sec:conclusion}
This work revisited the interference effect in an OFA based subnet population during training. Interference was believed to be the constraining factor in achieving good performance. To limit interference, OFA proposed progressive shrinking as a novel training method. However, we showed that interference is not the only constraining factor. Additionally, we must consider the subnet sampling bias of our selection scheme during training. Subnet sampling bias states that subnets which receive the most training will perform the best. We drew this conclusion by first connecting the subnet sampling bias to the data imbalance problem in general classification training. We then demonstrate this by proposing a simple-yet-effective training method called Random Subnet Sampling (RSS), which only mitigates the sampling bias, not interference. By mitigating only the bias, RSS trains a better performing subnet population than progressive shrinking on four small-to-medium datasets while also being 1.9 times faster. Furthering our experiments provided insight into both the interference and bias problems. We found interference to be more significant when combing the gradients of multiple sampled subnets. Additionally, we found the impact of bias depends on the subnet training sequence. In the future, we plan to test our hypothesis in large computer vision datasets such as the ImageNet dataset.


\section*{Acknowledgement}
This research was partly supported by Sentient Vision Systems. Sentient Vision Systems is one of the leading Australian developers of computer vision and artificial intelligence software solutions for defence and civilian applications.

{\small
\bibliographystyle{ieee_fullname}
\bibliography{MainPaper}

\begin{thebibliography}{10}\itemsep=-1pt

\bibitem{baker_designing_2017}
Bowen Baker, Otkrist Gupta, Nikhil Naik, and Ramesh Raskar.
\newblock Designing neural network architectures using reinforcement learning.
\newblock In {\em International Conference on Learning Representations}, volume
  abs/1611.02167, 2017.

\bibitem{bender_understanding_2018}
Gabriel Bender, Pieter-Jan Kindermans, Barret Zoph, Vijay Vasudevan, and Quoc
  Le.
\newblock Understanding and {Simplifying} {One}-{Shot} {Architecture} {Search}.
\newblock In {\em International {Conference} on {Machine} {Learning}}, pages
  550--559. PMLR, July 2018.
\newblock ISSN: 2640-3498.

\bibitem{brock_smash_2017}
Andrew Brock, Theodore Lim, James~M. Ritchie, and Nick Weston.
\newblock Smash: One-shot model architecture search through hypernetworks.
\newblock In {\em International Conference on Learning Representations}, volume
  abs/1708.05344, 2018.

\bibitem{cai_once-for-all_2020}
Han Cai, Chuang Gan, Tianzhe Wang, Zhekai Zhang, and Song Han.
\newblock Once for all: Train one network and specialize it for efficient
  deployment.
\newblock In {\em International Conference on Learning Representations}, 2020.

\bibitem{cai_proxylessnas_2019}
Han Cai, Ligeng Zhu, and Song Han.
\newblock Proxyless{NAS}: Direct neural architecture search on target task and
  hardware.
\newblock In {\em International Conference on Learning Representations}, 2019.

\bibitem{chen_net2net_2016}
Tianqi Chen, Ian~J. Goodfellow, and Jonathon Shlens.
\newblock Net2net: Accelerating learning via knowledge transfer.
\newblock In {\em International Conference on Learning Representations}, volume
  abs/1511.05641, 2016.

\bibitem{Chu_fairnas_2021}
Xiangxiang Chu, Bo Zhang, and Ruijun Xu.
\newblock Fairnas: Rethinking evaluation fairness of weight sharing neural
  architecture search.
\newblock In {\em Proceedings of the IEEE/CVF International Conference on
  Computer Vision (ICCV)}, pages 12239--12248, October 2021.

\bibitem{dai_fbnetv3_2021}
Xiaoliang Dai, Alvin Wan, Peizhao Zhang, Bichen Wu, Zijian He, Zhen Wei, Kan
  Chen, Yuandong Tian, Matthew Yu, P{\'e}ter Vajda, and Joseph~E. Gonzalez.
\newblock Fbnetv3: Joint architecture-recipe search using predictor
  pretraining.
\newblock {\em 2021 IEEE/CVF Conference on Computer Vision and Pattern
  Recognition (CVPR)}, pages 16271--16280, 2021.

\bibitem{dong_focal_2020}
Jianxiang Dong.
\newblock Focal {Loss} {Improves} the {Model} {Performance} on {Multi}-{Label}
  {Image} {Classifications} with {Imbalanced} {Data}.
\newblock {\em Proceedings of the 2nd International Conference on Industrial
  Control Network And System Engineering Research}, 2020.

\bibitem{estabrooks_multiple_2004}
Andrew Estabrooks.
\newblock A multiple resampling method for learning from imbalanced data sets.
\newblock {\em Computational Intelligence}, pages 18--36, 2004.

\bibitem{guo_single_2020}
Zichao Guo, Xiangyu Zhang, Haoyuan Mu, Wen Heng, Zechun Liu, Yichen Wei, and
  Jian Sun.
\newblock Single path one-shot neural architecture search with uniform
  sampling.
\newblock In {\em ECCV}, 2020.

\bibitem{ha_hypernetworks_2016}
David Ha, Andrew~M. Dai, and Quoc~V. Le.
\newblock Hypernetworks.
\newblock In {\em International Conference on Learning Representations}, volume
  abs/1609.09106, 2017.

\bibitem{han_learning_2015}
Song Han, Jeff Pool, John Tran, and William~J. Dally.
\newblock Learning both weights and connections for efficient neural networks.
\newblock In {\em Proceedings of the 28th International Conference on Neural
  Information Processing Systems - Volume 1}, NIPS'15, page 1135–1143,
  Cambridge, MA, USA, 2015. MIT Press.

\bibitem{howard_searching_2019}
Andrew~G. Howard, Mark Sandler, Grace Chu, Liang-Chieh Chen, Bo Chen, Mingxing
  Tan, Weijun Wang, Yukun Zhu, Ruoming Pang, Vijay Vasudevan, Quoc~V. Le, and
  Hartwig Adam.
\newblock Searching for mobilenetv3.
\newblock {\em 2019 IEEE/CVF International Conference on Computer Vision
  (ICCV)}, pages 1314--1324, 2019.

\bibitem{hu_dsnas_2020}
Shou-Yong Hu, Sirui Xie, Hehui Zheng, Chunxiao Liu, Jianping Shi, Xunying Liu,
  and Dahua Lin.
\newblock Dsnas: Direct neural architecture search without parameter
  retraining.
\newblock {\em 2020 IEEE/CVF Conference on Computer Vision and Pattern
  Recognition (CVPR)}, pages 12081--12089, 2020.

\bibitem{huang_ponas_2020}
Sian-Yao Huang and Wei-Ta Chu.
\newblock Ponas: Progressive one-shot neural architecture search for very
  efficient deployment.
\newblock {\em 2021 International Joint Conference on Neural Networks (IJCNN)},
  pages 1--9, 2021.

\bibitem{krizhevsky_learning_2009}
Alex Krizhevsky.
\newblock Learning multiple layers of features from tiny images.
\newblock 2009.

\bibitem{laurikkala_improving_2001}
Jorma Laurikkala.
\newblock Improving identification of difficult small classes by balancing
  class distribution.
\newblock In {\em AIME}, 2001.

\bibitem{lecun_gradient_1998}
Y. Lecun, L. Bottou, Y. Bengio, and P. Haffner.
\newblock Gradient-based learning applied to document recognition.
\newblock {\em Proceedings of the IEEE}, 86(11):2278--2324, 1998.

\bibitem{lecun_mnist_2010}
Yann LeCun, Corinna Cortes, and CJ Burges.
\newblock Mnist handwritten digit database.
\newblock {\em ATT Labs [Online]. Available: http://yann.lecun.com/exdb/mnist},
  2, 2010.

\bibitem{lin_focal_2018}
Tsung-Yi Lin, Priya Goyal, Ross~B. Girshick, Kaiming He, and Piotr Doll{\'a}r.
\newblock Focal loss for dense object detection.
\newblock {\em IEEE Transactions on Pattern Analysis and Machine Intelligence},
  42:318--327, 2020.

\bibitem{liu_progressive_2018}
Chenxi Liu, Barret Zoph, Jonathon Shlens, Wei Hua, Li-Jia Li, Li Fei-Fei,
  Alan~Loddon Yuille, Jonathan Huang, and Kevin~P. Murphy.
\newblock Progressive neural architecture search.
\newblock In {\em ECCV}, 2018.

\bibitem{liu_darts_2019}
Hanxiao Liu, Karen Simonyan, and Yiming Yang.
\newblock {DARTS}: Differentiable architecture search.
\newblock In {\em International Conference on Learning Representations}, 2019.

\bibitem{loshchilov_online_2016}
Ilya Loshchilov and Frank Hutter.
\newblock Online batch selection for faster training of neural networks.
\newblock In {\em International Conference on Learning Representations
  Workshop}, volume abs/1511.06343, 2015.

\bibitem{pham_efficient_2018}
Hieu Pham, Melody~Y. Guan, Barret Zoph, Quoc~V. Le, and Jeff Dean.
\newblock Efficient {Neural} {Architecture} {Search} via {Parameter} {Sharing}.
\newblock {\em arXiv:1802.03268 [cs, stat]}, Feb. 2018.
\newblock arXiv: 1802.03268.

\bibitem{real_large-scale_2017}
Esteban Real, Sherry Moore, Andrew Selle, Saurabh Saxena, Yutaka~Leon Suematsu,
  Jie Tan, Quoc~V. Le, and Alexey Kurakin.
\newblock Large-scale evolution of image classifiers.
\newblock In {\em Proceedings of the 34th International Conference on Machine
  Learning - Volume 70}, ICML'17, page 2902–2911. JMLR.org, 2017.

\bibitem{sahni_compofa_2021}
Manas Sahni, Shreya Varshini, Alind Khare, and Alexey Tumanov.
\newblock {C}omp{OFA}: Compound once-for-all networks for faster multi-platform
  deployment.
\newblock In {\em Proc. of the 9th International Conference on Learning
  Representations}, ICLR '21, May 2021.

\bibitem{shrivastava_training_2016}
Abhinav Shrivastava, Abhinav~Kumar Gupta, and Ross~B. Girshick.
\newblock Training region-based object detectors with online hard example
  mining.
\newblock {\em 2016 IEEE Conference on Computer Vision and Pattern Recognition
  (CVPR)}, pages 761--769, 2016.

\bibitem{simo-serra_fracking_2015}
Edgar Simo-Serra, Eduard Trulls, Luis Ferraz, Iasonas Kokkinos, and Francesc
  Moreno-Noguer.
\newblock Fracking {Deep} {Convolutional} {Image} {Descriptors}.
\newblock {\em arXiv:1412.6537 [cs]}, Feb. 2015.
\newblock arXiv: 1412.6537.

\bibitem{wang_attentivenas_2021}
Dilin Wang, Meng Li, Chengyue Gong, and Vikas Chandra.
\newblock Attentivenas: Improving neural architecture search via attentive
  sampling.
\newblock {\em 2021 IEEE/CVF Conference on Computer Vision and Pattern
  Recognition (CVPR)}, pages 6414--6423, 2021.

\bibitem{wang_unsupervised_2015}
Xiaolong Wang and Abhinav Gupta.
\newblock Unsupervised learning of visual representations using videos.
\newblock In {\em Proceedings of the IEEE international conference on computer
  vision}, pages 2794--2802, 2015.

\bibitem{weiss_effect_2001}
Gary~M. Weiss and Foster~J. Provost.
\newblock The effect of class distribution on classifier learning.
\newblock 2001.

\bibitem{wu_fbnet_2019}
Bichen Wu, Xiaoliang Dai, Peizhao Zhang, Yanghan Wang, Fei Sun, Yiming Wu,
  Yuandong Tian, P{\'e}ter Vajda, Yangqing Jia, and Kurt Keutzer.
\newblock Fbnet: Hardware-aware efficient convnet design via differentiable
  neural architecture search.
\newblock {\em 2019 IEEE/CVF Conference on Computer Vision and Pattern
  Recognition (CVPR)}, pages 10726--10734, 2019.

\bibitem{xia_progressive_2021}
Xin Xia, Xuefeng Xiao, Xing Wang, and Min Zheng.
\newblock Progressive automatic design of search space for one-shot neural
  architecture search.
\newblock In {\em 2022 IEEE/CVF Winter Conference on Applications of Computer
  Vision (WACV)}, pages 3525--3534, 2022.

\bibitem{xiao_fashion-mnist_2017}
Han Xiao, Kashif Rasul, and Roland Vollgraf.
\newblock Fashion-mnist: a novel image dataset for benchmarking machine
  learning algorithms.
\newblock {\em CoRR}, abs/1708.07747, 2017.

\bibitem{xu_analyzing_2021}
Jin Xu, Xu Tan, Kaitao Song, Renqian Luo, Yichong Leng, Tao Qin, Tie-Yan Liu,
  and Jian Li.
\newblock Analyzing and {Mitigating} {Interference} in {Neural} {Architecture}
  {Search}.
\newblock {\em arXiv:2108.12821 [cs]}, Aug. 2021.
\newblock arXiv: 2108.12821.

\bibitem{xu_scaling_2018}
Xiaowei Xu, Yukun Ding, Sharon Hu, Michael~Thaddeus Niemier, Jason Cong, Yu Hu,
  and Yiyu Shi.
\newblock Scaling for edge inference of deep neural networks.
\newblock {\em Nature Electronics}, 1:216--222, 2018.

\bibitem{yu_bignas_2020}
Jiahui Yu, Pengchong Jin, Hanxiao Liu, Gabriel Bender, Pieter-Jan Kindermans,
  Mingxing Tan, Thomas Huang, Xiaodan Song, and Quoc~V. Le.
\newblock Bignas: Scaling up neural architecture search with big single-stage
  models.
\newblock In {\em ECCV}, 2020.

\bibitem{yu_slimmable_2018}
Jiahui Yu, Linjie Yang, Ning Xu, Jianchao Yang, and Thomas Huang.
\newblock Slimmable neural networks.
\newblock In {\em International Conference on Learning Representations}, 2019.

\bibitem{zoph_neural_2017}
Barret Zoph and Quoc~V. Le.
\newblock Neural architecture search with reinforcement learning.
\newblock In {\em International Conference on Learning Representations}, volume
  abs/1611.01578, 2017.

\end{thebibliography}


\begin{thebibliography}{1}\itemsep=-1pt

\bibitem{cai_proxylessnas_2019}
Han Cai, Ligeng Zhu, and Song Han.
\newblock Proxyless{NAS}: Direct neural architecture search on target task and
  hardware.
\newblock In {\em International Conference on Learning Representations}, 2019.

\bibitem{howard_searching_2019}
Andrew~G. Howard, Mark Sandler, Grace Chu, Liang-Chieh Chen, Bo Chen, Mingxing
  Tan, Weijun Wang, Yukun Zhu, Ruoming Pang, Vijay Vasudevan, Quoc~V. Le, and
  Hartwig Adam.
\newblock Searching for mobilenetv3.
\newblock {\em 2019 IEEE/CVF International Conference on Computer Vision
  (ICCV)}, pages 1314--1324, 2019.

\bibitem{krizhevsky_learning_2009}
Alex Krizhevsky.
\newblock Learning multiple layers of features from tiny images.
\newblock 2009.

\end{thebibliography}
}

\end{document}


\title{Supplementary Material for the Paper: Does Interference Exist When Training a Once-For-All Network?}  


%
%
\maketitle
\thispagestyle{empty}

\section{RSS-Short Ablation Results}

We conducted additional ablation experiments to study the effect of the number of subnets sampled and the selection scheme on the proposed RSS-Short method. As mentioned earlier, RSS trains for 590 epochs and RSS-Short trains for 180 epochs. Other than this, all RSS-Short experiments shared the same hyperparameters as their RSS counterparts. All experiments are conducted on the CIFAR100 dataset~\cite{krizhevsky_learning_2009}.

\subsection{Effect of Number of Subnets Sampled}
\begin{figure}[b]
\centering
	\includegraphics[scale=0.55]{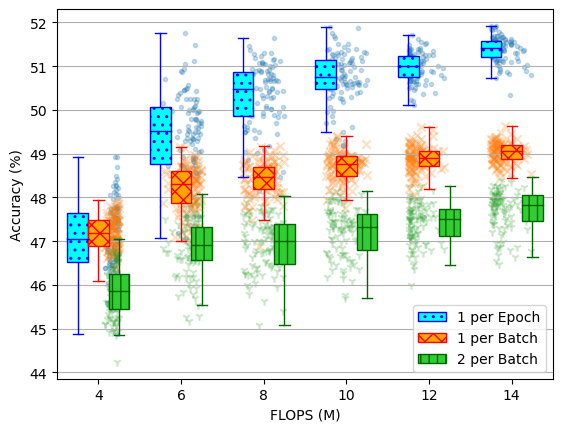}
	\caption{RSS-Short (proposed) results for sampling different numbers of subnets during training. The same pattern can be observed as in the main ablation results with per batch sampling and combined gradients performing worse than per epoch sampling.}
	\label{fig:sup_sample_change}
\end{figure}

\cref{fig:sup_sample_change} shows the results from comparing per epoch to per batch sampling during training. These results follow the results presented in the paper with a decrease in population accuracy between per epoch and per batch methods. A further drop occurs when training two subnets per batch and combining their gradients.

\begin{figure}[b]
\centering
	\includegraphics[scale=0.55]{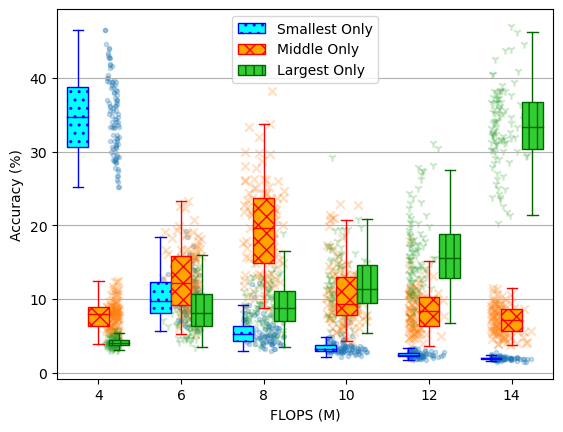}
	\caption{RSS-Short (proposed) population results from training a single subnet only. The results are the same as in the main ablation results.}
	\label{fig:sup_anchor}
\end{figure}

\subsection{Effect of Different Selection Schemes}
\begin{figure}[t]
\centering
	\includegraphics[scale=0.5]{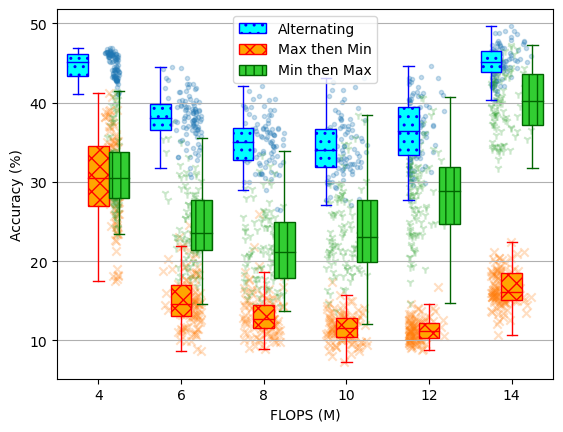}
	\caption{RSS-Short (proposed) population results from training the same largest and smallest subnets in different sequences. These results show a more significant bias towards the largest and smallest subnets than the main ablation results.}
	\label{fig:sup_anchor_seq}
\end{figure}

\cref{fig:sup_anchor,fig:sup_anchor_seq} show the results from altering the subnet selection scheme during training. In \cref{fig:sup_anchor}, the selection schemes tested are \textbf{Smallest Only}, \textbf{Middle Only} and \textbf{Largest Only}, where the smallest, middle and largest subnets train for the entire 180 epochs. These results again follow the results presented in the paper. Each single subnet selection scheme shows a heavy bias towards the trained subnet.

\cref{fig:sup_anchor_seq} shows the two subnet selection scheme with different training orders. The two subnets trained are the largest (\textbf{max}) and smallest (\textbf{min}). \textbf{Max then Min} trains the largest subnet for the first 90 epochs and the smallest subnet for the remaining 90 epochs. \textbf{Min then Max} does the opposite, training the smallest then the largest. \textbf{Alternating} switches between the two subnets, training each for one epoch at a time. The results in \cref{fig:sup_anchor_seq} differ slightly from the RSS results presented in the paper. For \textbf{Max then Min}, min has better performance than max; for \textbf{Min then Max}, max has better performance than min; and for \textbf{Alternating}, min and max have similar performance.  The accuracy of subnets in the 8-10 MFLOP range trained via the \textbf{Alternating} selection scheme perform significantly worse than smaller and larger subnets. This result is not seen in the main results, suggesting that additional epochs are required to reduce the bias in the subnet selection scheme.

\begin{figure}
\centering
	\begin{subfigure}[Figure A]{\columnwidth}
	\centering
		\includegraphics[scale=0.5]{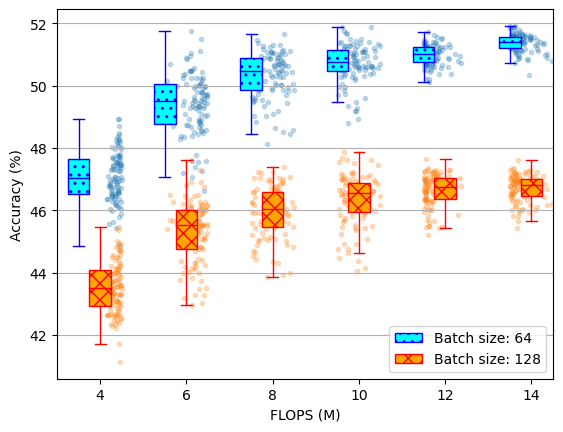}
		\caption{Resulting RSS-Short (proposed) subnet populations.}
	\end{subfigure}
	\begin{subfigure}[Figure B]{\columnwidth}
	\centering
		\includegraphics[scale=0.5]{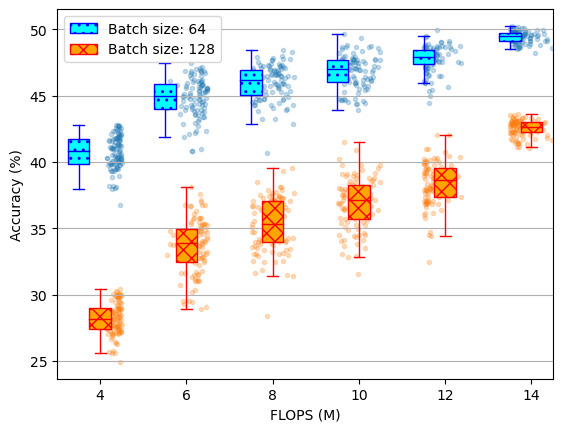}
		\caption{Resulting OFA subnet populations.}
	\end{subfigure}
	\caption{Resulting RSS-Short (proposed) and OFA subnet populations from increasing the batch size during training. These results show that the batch size has the same effect for both methods.}
	\label{fig:sup_batch_change}
\end{figure}

\section{Hyperparameter Ablation Studies}
\begin{figure*}[h]
	\centering
	\begin{subfigure}[Figure A]{\columnwidth}
	\centering
		\includegraphics[scale=0.5]{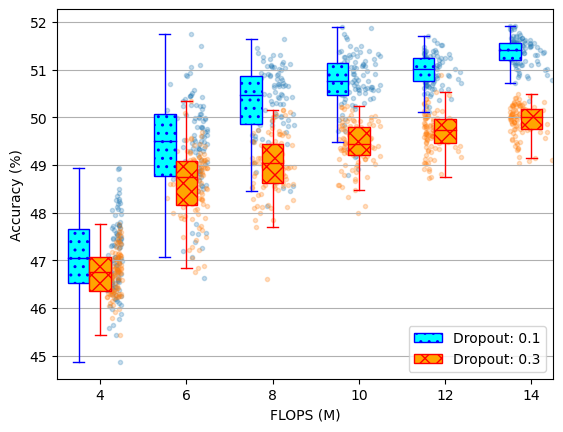}
		\caption{Resulting RSS-Short (proposed) subnet populations.}
	\end{subfigure}
	\begin{subfigure}[Figure B]{\columnwidth}
	\centering
		\includegraphics[scale=0.5]{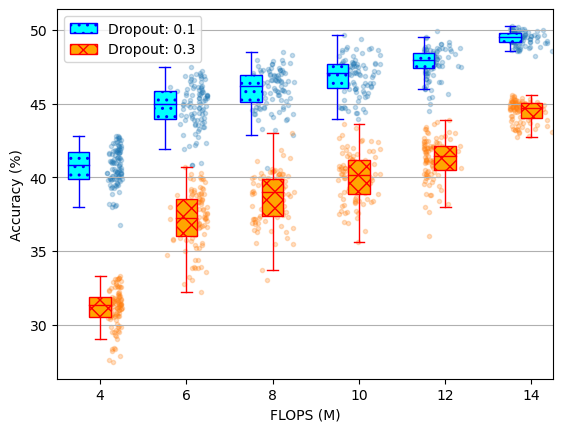}
		\caption{Resulting OFA subnet populations.}
	\end{subfigure}
	\caption{Resulting RSS-Short (proposed) and OFA subnet populations from increasing the drop-out rate. These results show that both methods suffer from an increased dropout rate with OFA having a more significant accuracy drop.}
	\label{fig:sup_dropout_change}
\end{figure*}

\begin{figure*}[h]
\centering
	\begin{subfigure}[Figure A]{\columnwidth}
	\centering
		\includegraphics[scale=0.5]{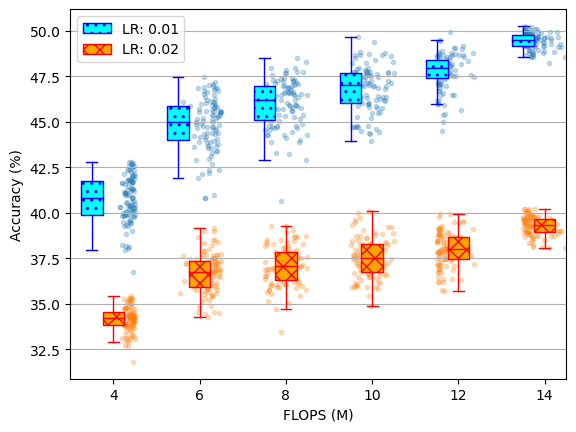}
		\caption{Resulting RSS-Short (proposed) subnet populations.}
	\end{subfigure}
	\begin{subfigure}[Figure B]{\columnwidth}
	\centering
		\includegraphics[scale=0.5]{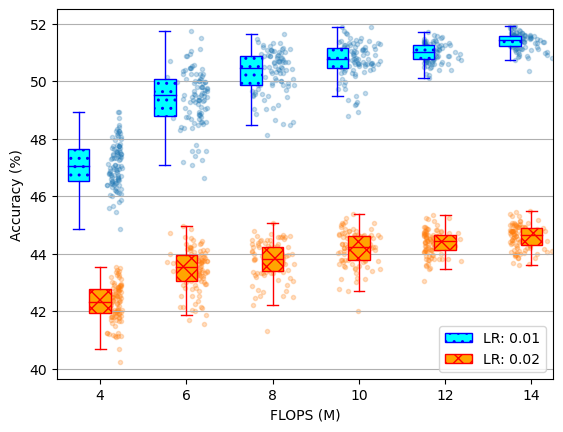}
		\caption{Resulting OFA subnet populations.}
	\end{subfigure}
	\caption{Resulting RSS-Short (proposed) and OFA subnet populations from increasing the learning rate. These results show that the learning rate decreases the accuracy of both methods.}
	\label{fig:sup_lr_change}
\end{figure*}

We alter various hyperparameters to ensure our findings are not limited to specific hyperparameter values. We adjust the batch size, dropout rate and learning rate, comparing the effect on RSS-Short and OFA. \cref{fig:sup_batch_change} shows that increasing the batch size from 64 to 128 results in a overall accuracy drop for both RSS-Short and OFA. \cref{fig:sup_dropout_change} shows that increasing the dropout rate from 0.1 to 0.3 again results in a decrease in accuracy for both methods, with a more significant decrease for OFA. Lastly, \cref{fig:sup_lr_change} shows that increasing the learning rate from 0.01 to 0.02 reduces the accuracy of both methods. These results show that hyperparameter changes have the same effect for both methods. Confirming our results are not limited to specific hyperparameter values.

\section{ProxylessNAS Base Architecture}
\begin{figure*}[h]
	\centering
	\includegraphics[scale=0.5]{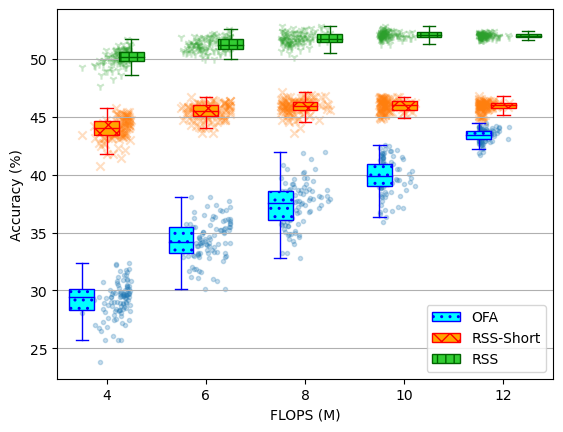}
	\caption{Resulting OFA, RSS-Short and RSS subnet populations from training with a ProxylessNAS~\cite{cai_proxylessnas_2019} base architecture on CIFAR100~\cite{krizhevsky_learning_2009}. These results are consistent with the main results despite a lower overall accuracy for each method than with the MobileNetV3~\cite{howard_searching_2019} base architecture.}
	\label{fig_sup_proxyless}
\end{figure*}

We change the base architecture used during training, showing that results obtained are not unique to the MobileNetV3~\cite{howard_searching_2019} base architecture. \cref{fig_sup_proxyless} shows this as we switch to using a ProxylessNAS~\cite{cai_proxylessnas_2019} base architecture and achieve similar results. The population settings remain the same as before; however, the resulting subnet population only ranges from 4 MFLOPs to 12 MFLOPs. Therefore, we only show subnets from sizes 4, 6, 8, 10 and 12 MFLOPs. 

\bibliographystyle{ieee_fullname}
\bibliography{supplementary}